\documentclass{article}

\usepackage{microtype}
\usepackage{graphicx}
\usepackage{subfigure}
\usepackage{booktabs} 

\usepackage{hyperref}

\usepackage[accepted]{icml2022}

\usepackage{amsmath}
\usepackage{amssymb}
\usepackage{mathtools}
\usepackage{amsthm}
\usepackage{csquotes}
\usepackage{subfiles}
\theoremstyle{plain}
\newtheorem{theorem}{Theorem}[section]

\newtheorem{lemma}[theorem]{Lemma}

\theoremstyle{definition}
\newtheorem{definition}[theorem]{Definition}

\theoremstyle{remark}

\usepackage{stackengine}

\makeatletter
\newcommand{\distas}[1]{\mathbin{\overset{#1}{\kern\z@\sim}}}%

\usepackage[textsize=tiny]{todonotes}

\icmltitlerunning{Architecture Agnostic Federated Learning for Neural Networks}

\begin{document}

\twocolumn[
\icmltitle{Architecture Agnostic Federated Learning for Neural Networks}

\icmlsetsymbol{equal}{*}

\author{
  Disha Makhija \\
  The University of Texas at Austin, \\
  \texttt{disham@utexas.edu}
   \And
   Nhat Ho \\
   The University of Texas at Austin, \\
   \texttt{minhnhat@utexas.edu}
   \And 
   Joydeep Ghosh \\
   University of Texas at Austin, \\
   \texttt{jghosh@utexas.edu}
}

\begin{icmlauthorlist}
\icmlauthor{Disha Makhija}{ut}
\icmlauthor{Xing Han}{ut}
\icmlauthor{Nhat Ho}{ut}
\icmlauthor{Joydeep Ghosh}{ut}
\end{icmlauthorlist}

\icmlaffiliation{ut}{The University of Texas at Austin, Austin, Texas, USA}

\icmlcorrespondingauthor{Disha Makhija}{disham@utexas.edu}

\icmlkeywords{Federated Learning, Machine Learning, Distributed Learning, Heterogeneous Models}

\vskip 0.3in
]
\printAffiliationsAndNotice{}


\begin{abstract}
With growing concerns regarding data privacy and rapid increase in data volume, Federated Learning (FL) has become an important learning paradigm. However, jointly learning a deep neural network model in a FL setting proves to be a non-trivial task because of the complexities associated with the neural networks, such as varied architectures across clients, permutation invariance of the neurons, and presence of non-linear transformations in each layer. This work introduces a novel framework, \emph{Federated Heterogeneous Neural Networks} (FedHeNN), that allows each client to build a personalised model without enforcing a common architecture across clients. This allows each client to optimize with respect to local data and compute constraints, while still benefiting from the learnings of other (potentially more powerful) clients. The key idea of FedHeNN is to use the instance-level representations obtained from peer clients to guide the simultaneous training on each client. The extensive experimental results demonstrate that the FedHeNN framework is capable of learning better performing models on clients in both the settings of homogeneous and heterogeneous architectures across clients.


\end{abstract}

\section{Introduction}
Distributed machine learning has been an important field of study for long and is becoming more and more important with time~\cite{paka03}. Federated Learning is a type of distributed machine learning setting that consists of many end devices or silo organisations (\textit{clients}) which have access to the data stored locally and a global server which can orchestrate the learning without accessing all the data. With the ever so rapidly growing amount of data and the concerns around data privacy, federated learning has emerged as a very promising direction, as it allows learning of a global model using the data present at each client but without explicitly sharing the data outside the client devices, thus helping in ensuring data privacy and also reducing the cost of centralised training and storage. 

FedAvg~\cite{pmlr-v54-mcmahan17a} is the de-facto federated learning algorithm where each client performs SGD steps towards training its local model using its own data and compute resources. The client models are then periodically shared with the server and the server aggregates the client models to create a global model which is sent back to the clients. However, the solution obtained from FedAvg has been shown to diverge in presence of statistical heterogeneity across clients~\cite{fedprox}. Over the years several modifications have been proposed to the original algorithm addressing different aspects like data heterogeneity, availability of clients for training, modifying the aggregation mechanism, optimizing communication costs, personalised client models etc.~\cite{fedprox},~\cite{ditto},~\cite{scaffold},~\cite{semicyclic},~\cite{mocha},~\cite{fedma},~\cite{fedrep}. In all of these methods the global model parameters are obtained by appropriately aggregating the local model parameters.  

\begin{figure}[ht]
\vskip 0.2in
\begin{center}
\centerline{\includegraphics[width=\columnwidth, height =6cm]{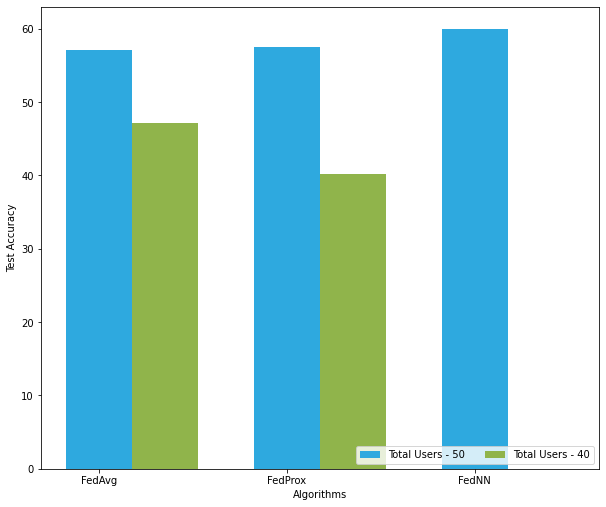}}
\caption{When a fraction of clients cannot afford to build full-blown models, FedAvg and FedProx need to drop the clients altogether resulting in poorer average performance versus FedHeNN which can accommodate clients with heterogeneous architectures.}
\label{change_num_users}
\end{center}
\vskip -0.2in
\end{figure}

Yet in most practical settings where the clients are heterogeneous and differ a lot in their compute resources and data distributions, these methods may face significant challenges. Several real-world FL scenarios require training over end devices which have very different hardware. In such cases, for the above methods, the clients that are incapable of training large models will never be able to take part in the training process. Similarly, if the common model architecture is kept small to accommodate all the clients, some clients will be under-utilised. These negative effects might become more prominent in the cross-silo FL setting where the total number of clients is even smaller (typically less than 100). An illustrative experiment shown in Figure \ref{change_num_users} demonstrates the drop in accuracy of the global methods like FedAvg and FedProx when a few clients are left out of the training.

In this work, we propose a systematic framework for architecture agnostic federated learning called FedHeNN. This framework is able to overcome the aforementioned challenges of client heterogeneity by allowing each client to build a personalised model without any constraint on the model architecture including number and size of the hidden layers, activation functions, etc. The learning across clients is transferred by grounding the representations being learnt at each client through a proximal term. Specifically, the optimisation objective at each client comprises of two terms - the task loss, and a proximal term that pulls the final learned representations of the models together. This also makes the neural networks learn more robust representations on a wide range of data leading to better performance. We test our framework on a suite of federated learning datasets in both the settings of homogeneous and heterogeneous model architectures across clients. 

\textbf{Our Contributions} are summarized as follows :
\vspace{-0.1in}
\begin{enumerate}
    \item Our primary contribution is FedHeNN, a new framework for training deep neural networks in federated learning settings. We identify the shortcomings in joint learning of neural-networks and circumvent those by grounding the representations being learnt at each client through a proximal term. We empirically show that having the proximal term on the representations can deliver superior performance.
    \item To allow transfer of learning across architectures in different output spaces, we suggest the use of a kernel based distance metric called Centered Kernel Alignment (CKA). The use of kernel based distance metric provides FedHeNN greater flexibility.
    \item Additionally, the structure of FedHeNN allows itself to be extended to the setting where different clients have different model architectures. Thus, we propose a solution that is architecture agnostic and has performance that is better or comparable to the existing methods that operate in homogeneous architecture settings.
\end{enumerate}

The rest of the paper is organised as follows. Section~\ref{lit_review} provides a background on Federated Learning and related developments. In Section~\ref{methodology}, we go over the preliminaries and then propose our framework, FedHeNN. We provide a thorough experimental evaluation of FedHeNN on different types of FL datasets in Section~\ref{experiments} and conclude in Section~\ref{discussion}.  

\section{Related Work} \label{lit_review}
Distributed learning algorithms were extensively studied in the data mining community in the early 2000s under topics such as distributed (and privacy preserving) data mining~\cite{lipi00},~\cite{megh03},~\cite{megh05},~\cite{paka03},~\cite{ghtu00}. This topic was re-framed as \enquote{Federated Learning} in an influential paper~\cite{pmlr-v54-mcmahan17a} that introduced the FedAvg algorithm, and has rapidly gained new adherents since then.

The framework proposed by FedAvg~\cite{pmlr-v54-mcmahan17a} has been the standard solution for Federated Learning since 2017. However when the data across clients is non-iid, averaging the local optima of the multiple clients to obtain the global solution may lead to divergence in the optimization. To solve this, FedProx~\cite{fedprox} proposes to modify the local training objective by adding a proximal term which penalizes the distance between the current global model weights and the local model weights thus preventing each local update from moving far away. Similarly, SCAFFOLD~\cite{scaffold} introduces control-variates to correct the local updates. FedPD~\cite{fedpd}, FedSplit~\cite{fedsplit}, and FedDyn~\cite{feddyn} are other important works that study the problem of finding better fixed points.

\cite{fedDF} shows that simply averaging the local models learnt on local distributions to obtain a global model may not be ideal. FedBE~\cite{fedbe} also provides evidence that the best performing aggregate model need not necessarily be the average of the local models. Another important aspect of Federated Learning is thus to appropriately aggregate the local models. PFNM~\cite{pfnm} and FedMA~\cite{fedma} show that neurons in each layer are permutation invariant and these works perform a layer by layer matching of neurons and then aggregate of the matched neurons. ~\cite{singh2020model} use an Optimal-Transport based distance to perform the matching of neurons before aggregation. Different from above, FedDF~\cite{fedDF} uses additional data samples to distill knowledge from clients' local models. FedNova~\cite{fednova} allows each client to perform variable amounts of work and calculates a regularized average for the aggregate.

Several works focus on creating local personalized models instead of a single global model to cater to the heterogeneity of the data distributions across clients. In the literature, Personalised FL has been solved using various different ways like keeping the task specific component of the clients local~\cite{fedrep}, clustering the clients~\cite{SattlerMS21},~\cite{GhoshCYR20}, using meta-learning ~\cite{personalised_meta_learning},~\cite{waffle},~\cite{maml},~\cite{KhodakBT19}, multi-task learning~\cite{fed_mtl},~\cite{ditto}, ~\cite{mocha} and transfer-learning~\cite{local_adaption}. 

\cite{HaoELZLCC21},~\cite{FLviaSyn},~\cite{luo2021fear} augment the data distribution by creating additional data samples and using them for learning. MOON~\cite{li2021model} uses a contrastive loss to bring the representations of objects closer and utilises it to correct the local training.

The need for heterogeneous client models was recently highlighted in FedProto~\cite{tan2021fedproto}. Our work is different from them as our formulation directly works on representations of data instances as opposed to class specific prototypes  being used in FedProto for learning and inference. We think that having only one prototype per class could be limiting in case of multi-modal classes.

\section{FedHeNN Methodology} \label{methodology}
In an effort to enhance FL with heterogeneous clients, we propose a new framework, \emph{Federated Heterogeneous Neural Networks} (FedHeNN). FedHeNN provides a systematic way to achieve the goal of joint learning of deep neural networks varying in architecture and output space across distributed clients. In this section, we will first cover the preliminaries and then go to the proposed method. The algorithms for homogeneous and heterogeneous settings are  described in Algorithm~\ref{alg:fednn_algo_homo} and~\ref{alg:fednn_algo_hetero} respectively.

\textbf{Preliminaries} We first go over the key concepts of the federated learning methods and then elaborate on our proposed method. 
Consider a set of clients {$i \in |N|$}, traditional FL algorithm learns a single global model $\hat{\mathcal{W}}$ by using the model parameters learned on individual clients: $\hat{\mathcal{W}} =  g(\mathcal{W}_1,\mathcal{W}_2,\dots,\mathcal{W}_N),$ 
{where $\mathcal{W}_i$ are the local parameters obtained at the $i^{th}$ client and $g$ is the  aggregation function}. For a single-layer model, $\mathcal{W} = $ \textbf{w} is the vector of weights. For an $m$-layer model $\mathcal{W} = (W^1, W^2,\dots,W^m)$ is the collection containing weights at each layer. We assume the bias terms are incorporated in the weights. 

\subsection{FedAvg}
FedAvg learns each client's local parameters by solving
\vspace{-0.08cm}
\begin{equation}
    \min_{\mathcal{W}_i} \mathcal{L}_i = \mathcal{F}({\mathcal{W}_i}) = \dfrac{1}{n_i}\sum_{j=1}^{n_i}  \ell(y_j, \hat{y}_j ; \mathcal{W}_i).
\end{equation}
where $\ell(.)$ is the loss function. An element-wise average is then performed on the weight matrices of the clients for getting an aggregated model $\hat{\mathcal{W}}$ at the server. In case of a multi-layer network, the average is taken layer-wise and the parameters of each client are weighted based on the number of samples present at the client. For the $l^{th}$ layer, we have
$$
\hat{\mathcal{W}^l} = \dfrac{1}{\sum_{k} n_k} \sum_{i=1}^{N} n_i \mathcal{W}^l_i.
$$
While this aggregation mechanism is shown to have good empirical results, this method of learning involving element-wise averaging might give sub-optimal results because of various reasons like permutation invariance property of the neurons, different data distributions across clients etc. Besides, FedAvg imposes a hard constraint on clients to train the models with the exact same architectures. In practice, it is highly likely that different clients may not be able to train the models of same capacity.

We identify that apart from solving for sub-problems like aggregation mechanism or data heterogeneity, we also need to make sure that the networks on clients are being trained towards the same goal. In order to do so we suggest to modify the objective function being optimized at each client. 

\begin{figure}[t!]
\vskip 0.2in
\begin{center}
\centerline{\includegraphics[scale=0.5]{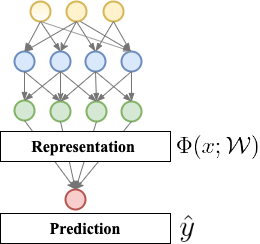} }
\caption{Depiction of a Neural Network consisting of two components - representation learning part and prediction function.}
\label{fednn_rep}
\end{center}
\vskip -0.2in
\end{figure}

We consider a neural network to be composed of two components - the representation learning component that maps the input to a $k$-dimensional representation vector and the task learning component which learns the prediction function. The initial layers of the neural network are considered as the representation learning component whose output is the representation vector denoted by $\Phi(x;\mathcal{W})$ for a data instance $x$ under the model with parameters $\mathcal{W}$, while the last layer is considered to be the task-specific layer and the output of which is the prediction corresponding to data instances denoted by $\hat{y}$, as depicted in Figure \ref{fednn_rep}. The details of the method are explained in the next sub-sections. Additionally, we explore two different settings for the FedHeNN framework - homogeneous and heterogeneous setting, which we define below and elaborate on in the next sub-section. 

\begin{definition}
\label{def:FL Settings}
We refer to a FL setting as \textbf{homogeneous} when each of the client in the network has the exact same architecture. A FL setting is said to be \textbf{heterogeneous} when each client in the network can define its own architecture.
\end{definition}

\subsection{FedHeNN for Homogeneous Clients}
For the homogeneous federated learning setting, we propose to modify the loss function on each client to additionally incorporate a proximal term on the representations. In particular, instead of just minimising the loss function $\mathcal{F}({\mathcal{W}_i})$, each client minimises the following objective -
\begin{equation}\label{eqn:homo1}
\begin{multlined}
\min_{\mathcal{W}_i} \mathcal{L}_i = \\
\mathcal{F}({\mathcal{W}_i}) + \eta \hspace{0.03in}  \text{d}(\Phi_i(\textbf{X} ; \mathcal{W}_i),\Phi_{\text{global}}(\textbf{X} ;\hat{\mathcal{W}}(t-1))).
\end{multlined}
\end{equation}

where $\text{d}(.,.)$ is some distance metric, $\Phi_i(\textbf{X} ; \mathcal{W}_i)$ is the representation learned on the $i^{th}$ client for the set of data points $\textbf{X}$, and the $\Phi_{\text{global}}(\textbf{X}; \hat{\mathcal{W}}(t-1))$ is the representation learned by the global model communicated at the previous round $t-1$. The $\hat{\mathcal{W}}$ is learned using the FedAvg algorithm. Because we are ensuring the similarities between the representations we do not explicitly need to address the aggregation mechanism. The proximal term helps in bringing the representations of the local and global model together and helps in correcting the learning on each client. 

\subsection{FedHeNN for Heterogeneous Clients}
In certain settings it might not be practical for all clients to train the neural network models of same complexity. We identify that it is non-trivial to use existing algorithms to perform joint training in such settings. 

Our framework allows one to perform federated learning across heterogeneous clients by collecting and aggregating the representations learnt on the local clients and pulling those representations closer. The framework is inspired by the same intuition that there is an underlying lower dimensional representation of the data that the clients can together uncover. To achieve this, we let each client train its own model but pull the representations learnt by different clients closer by adding a proximal term to the client's loss function. The proximal term measures the distance between the representations learnt by different local models. Specifically, if $\Phi_i(\textbf{X}; \mathcal{W}_i)$ is the representation matrix obtained on the $i^{th}$ client for data $\textbf{X}$, then the proximal term measuring the distances between client representations for the $i^{th}$ client can be written as -
$$
 \text{d} ( \Phi_i(\textbf{X}; \mathcal{W}_i), \Bar{\Phi}(\textbf{X} ; (t-1))   ),
$$
where
$$ \Bar{\Phi}(\textbf{X} ; (t-1)) =  \sum_{j=1}^{N} w_j \Phi_{j}(\textbf{X}; \mathcal{W}_j(t-1)).
$$
The contribution of each client $j$, $w_j$ could be set to reflect the capacity or the strength of the model on the $j^{th}$ client. At each iteration the server generates a set of unlabelled data instances called \textit{Representation Alignment Dataset} (RAD) denoted by $\textbf{X}$ and uses it to align the representations across clients. As before, the clients share the local weights with the server. Instead of aggregating the weights, the server uses each client's weights to generate representations for the RAD and aggregates the representations over clients. The server then distributes the aggregated representations and the RAD to each client. The clients on the next iteration learn a model that minimizes the loss that is a combination of the task loss and distance to the aggregated representations. 

Specifically, at iteration $t$ each client is trying to minimize 
\begin{equation}\label{eqn:hetero1}
\min_{\mathcal{W}_i} \mathcal{L}_i =
\mathcal{F}({\mathcal{W}_i}) + \eta \hspace{0.03in} \text{d} ( \Phi_i(\textbf{X}; \mathcal{W}_i),  \Bar{\Phi}(\textbf{X};(t-1))  ).
\end{equation}

This method helps us utilise the training from local clients but without explicitly generating a global aggregated model. The privacy of the clients data is also preserved as the clients do not need to share their local data. Thus the clients can keep on using the local personalised models but utilise the training from other peer clients.

Later when we describe the distance function d$(.,.)$ we will see that our method is robust to having output representations of clients in different spaces, i.e., $\Phi_i(x; \mathcal{W}_i)$'s, $\forall i$, need not be aligned with each other.


\subsection{The Proximal Term}
Here we give a detailed explanation about the proximal term or the distance function being used in our objective formulation in equations~\eqref{eqn:homo1} and \eqref{eqn:hetero1}.

As mentioned earlier, we consider a deep neural network model training to be performing two high level steps - learning to map the inputs of different modalities to a $k$-dimensional representation, and then learning a prediction function. We consider all but the last layer of the neural network as the representation learning module and its output as the learned representation and denote it by $\Phi(. ; .)$. So in order to match the representation learning part of the network across all clients we compare the outputs of the representation learning module of the networks on the same set of instances. In specific, we use a set of input examples ${x_l \in |L|}$ encoded in matrix $\textbf{X}$ a.k.a. RAD, pass them through all the local networks and capture the outputs of the representation learning module of the networks in matrices $A_i \in \mathbb{R}^{L \times d_i}$, where $A_i(l,:) = \Phi_i(x_l;\mathcal{W}_i)$. $A_i$ stands for the activation (representation) matrix of the $i^{th}$ client, $L$ is the size of RAD and $d_i$ is the output dimension of the representation. Note that our method can work with representations of different dimensions on different clients, i.e., $d_i$ can vary with $i$. At each communication round, we randomly select only a sample of the instances to be a part of the alignment dataset keeping $L << n_i,  \forall i$. 

After obtaining the activation matrices $A_i$'s, we need a way to compare the representations. For this, we suggest using the representation distance matrices (RDMs). An RDM uses the distance between the instances to capture the characteristics of the representational space. Representational distance learning~\cite{rdl} has also been used for knowledge distillation in the past.  

To measure the distance, we use a distance metric proposed specifically for neural network representations called Centered Kernel Alignment(CKA)~\cite{cka}. CKA was originally proposed to compare the representations obtained from different neural networks to determine equivalence relationships between hidden layers of neural networks. CKA takes into input the activation (representation) matrices and outputs a similarity score between 1 (identical representations) and 0 (not similar at all). Some of the useful properties of CKA include invariance to invertible linear transformation, orthogonal transformation and isotropic scaling. 

The other common distance metrics used in the literature for comparing representations from neural networks~\cite{cka, ding2021grounding} are based on Canonical Correlation Analysis(CCA) and Procrustes distance. The CCA based distance requires computing eigenvectors of a $n \times k$ matrix formed by $k$-dimensional representations of $n$ instances. But usage of eigenvector based function in the loss causes numerical instabilities while training through backpropagation thus making all CCA based distances unusable for training neural networks. The Procrustes based distance is only suitable for matrices with same dimensions and thus works only in the homogeneous settings. We did experiment with using orthogonal Procrustes distance for homogeneous settings in our method and found that the obtained performance is comparable with respect to CKA, for example, on CIFAR10, CKA obtains $94.7 \pm 1.1 \%$ accuracy and orthogonal Procrustes obtains $93.7 \pm 1.7 \%$. Also, the experiments shown in the paper that proposed CKA~\cite{cka} show the superiority of the CKA based method for comparing representations across different neural networks. Moreover, because CKA is a kernel based metric it provides a way to compare the representations obtained from networks of different widths thus providing greater flexibility for heterogeneous settings. The kernel based metric is able to characterize the representation space via the pairwise similarity between instances and thus helps in diminishing the effects of permutation invariance. CKA is also shown to be invariant to changes in initializations of the network. Because of all these properties, CKA extends itself naturally for use as a distance metric in our method.

Let $A_i \in \mathbb{R}^{L \times d_i}$ and $A_j \in \mathbb{R}^{L \times d_j}$ be the representation matrices for clients $i$ and $j$ obtained for the RAD. The distance between $A_i$ and $A_j$ is obtained by first computing the kernel matrices $K_i$ and $K_j$ for any choice of kernel $\mathcal{K}$:
$$K_i(p,q) = \mathcal{K}(A_i(p,:), A_i(q,:)), $$
$$K_j(p,q) = \mathcal{K}(A_j(p,:), A_j(q,:)), $$ 
where $K_i \in \mathbb{R}^{L \times L}$ and $K_j \in \mathbb{R}^{L \times L}$ are the representational distance matrices then similarity between $K_i$ and $K_j$ is given under CKA by -
$$
\text{CKA}(K_i,K_j) = \dfrac{\text{HSIC}(K_i,K_j)}{\sqrt{\text{HSIC}(K_i,K_i)\text{HSIC}(K_j,K_j)}}.
$$
where the estimator for Hilbert-Schmidt Independence Criterion (HSIC) could be written as -
$$ HSIC(K_i,K_j) = \dfrac{1}{(L-1)^2}\text{tr}(K_i H K_j H) \hspace{0.2in}. $$ \\ 
with $H$ as the centering matrix
$$\hspace{0.2in} H = I_L - \dfrac{1}{L}\textbf{1}\textbf{1}^T.$$ 

We use a linear and an RBF kernel for computing distances. The Linear CKA can be simply written as -
\begin{equation}\label{eqn:lin_cka}
\begin{multlined}
\text{Linear CKA}(K_i,K_j) = \\
\text{Linear CKA}(A_i A_i^T, A_j A_j^T) = \dfrac{||A_j^{T}A_i||^2_F}{||A_i^{T}A_i||_F ||A_j^{T}A_j||_F}. 
\end{multlined}
\end{equation}

The local learning objectives for our method in homogeneous and heterogeneous settings use CKA based dissimilarity and thus respectively become
\begin{equation}\label{eqn:homo2}
\begin{multlined}
\min_{\mathcal{W}_i} \mathcal{L}_i = 
\mathcal{F}({\mathcal{W}_i}) + \eta \hspace{0.03in} \text{d}_{\text{CKA}}(K_i, K_{\text{global}}(t-1)).
\end{multlined}
\end{equation}
where $K_{\text{global}}(t-1)$ is the representation distance matrix obtained over the instances from the global model at previous iteration, and
\begin{equation}\label{eqn:hetero2}
\begin{multlined}
\min_{\mathcal{W}_i} \mathcal{L}_i = 
\mathcal{F}({\mathcal{W}_i}) + \eta \hspace{0.03in} \text{d}_{\text{CKA}} \bigg( K_i, \Bar{K}(t-1) \bigg),
\end{multlined}
\end{equation}
where we define
$$ \Bar{K}(t-1)  =  \sum_{j=1}^{N} w_j K_j(t-1). $$

For $\eta = 0$, the objective function corresponding to homogeneous FedHeNN, in equation \eqref{eqn:homo2}, becomes the FedAvg algorithm. For the heterogeneous FedHeNN, the objective function given in equation \eqref{eqn:hetero2} boils down to individual clients training their own local models in isolation using only the local data. On the other hand, when $\eta \rightarrow \infty $, the framework will try to obtain identical representations from all the local models without caring about the prediction task. In homogeneous FL setting for FedHeNN with linear CKA, after sufficiently large number of iterations ${t}^\prime$, the framework will make all the models identical. The following lemma provides an intuitive explanation on the loss function.

\begin{lemma}
\label{lem:homo_lin_cka}
Given a homogeneous FL setting with clients training linear models and $\eta \rightarrow \infty$, then after sufficiently large number of iterations $t^\prime$, for the FedHeNN framework with Linear CKA, we have $\mathcal{W}_i = \hat{\mathcal{W}}$ for all $i$. 

\begin{proof}
The optimization problem at each client $i$ is
$$ \min_{\mathcal{W}_i} \mathcal{L}_i = 
\mathcal{F}({\mathcal{W}_i}) + \eta \hspace{0.03in} \text{d}_{\text{CKA}}(K_i, K_{\text{global}}(t-1)). $$
When $\eta \rightarrow \infty$, the problem becomes
$$ \max_{\mathcal{W}_i} \text{CKA}(K_i, K_{\text{global}}(t-1)). $$

For linear CKA, we have 
$$ \text{CKA}(K_i, K_{\text{global}}) = \dfrac{||A_{\text{global}}^{T} A_i||^2_F}{||A_i^{T} A_i||_F ||A_{\text{global}}^{T} A_{\text{global}}||_F}. $$ 

If we assume that each client has a linear network, then $A_i(\textbf{X}) \in \mathbb{R}^{L \times d_i} = \Phi_i(\textbf{X}; \mathcal{W}_i) = (\textbf{X} W_i^{(1)})$. 
Then, after sufficiently large number of iterations $t^\prime$, i.e., when each client has seen numerous RADs, optimizing equation \eqref{eqn:homo2} for $\eta \rightarrow \infty$ will lead to 
$$\mathcal{W}_i = \hat{\mathcal{W}} \quad \forall i.$$ 
As a consequence, we reach the conclusion of the lemma.
\end{proof}
\end{lemma}

\begin{algorithm}[tb]
   \caption{FedHeNN Algorithm for Homogeneous clients}
   \label{alg:fednn_algo_homo}
   \begin{algorithmic}
   \STATE {\bfseries Input:} number of clients $N$, number of communication rounds $T$, 
   number of local epochs $E$, parameter $\eta_0$ 
   \STATE {\bfseries Output:} Final model $\hat{\mathcal{W}}(T)$ \\
   
   \STATE {\bfseries At Server - }
   \STATE Initialize $\hat{\mathcal{W}}(0)$
   \FOR{$t=1$ {\bfseries to} $T$}
    \STATE $\eta = \eta_0 \times f(t)$
    \STATE Generate RAD, \textbf{X}, by random sampling
    \STATE Select a subset of clients $\mathcal{N}_t$ 
    \FOR{each selected client $i \in \mathcal{N}_t$}
    \STATE $\mathcal{W}_i(t) =$ \textbf{LocalTraining}$( \hat{\mathcal{W}}(t-1), \textbf{X}, \eta)$
   \ENDFOR
   \STATE $\hat{\mathcal{W}}(t) = \dfrac{1}{\sum_{j \in \mathcal{N}_t} n_j} \sum_{i \in \mathcal{N}_t} n_i \mathcal{W}_i(t)$
   \ENDFOR
   \STATE $\hat{\mathcal{W}}(T) = \dfrac{1}{\sum_{j \in |N|} n_j} \sum_{i \in |N|} n_i \mathcal{W}_i(t)$
   \STATE Return $\hat{\mathcal{W}}(T)$
   \STATE {\bfseries LocalTraining}$( \hat{\mathcal{W}}(t-1), \textbf{X}, \eta)$
   \STATE Initialize $\mathcal{W}_i(t)$ with $\hat{\mathcal{W}}(t-1)$
   \STATE  $A_{\text{global}}(\textbf{X}) = \Phi_{\text{global}}(\textbf{X};\hat{\mathcal{W}}(t-1))$
   \STATE  $K_{\text{global}}(t-1) = \mathcal{K}(A_{\text{global}}(\textbf{X}),A_{\text{global}}(\textbf{X}))$
   \FOR{each local epoch}
   \STATE  $A_i(\textbf{X}) = \Phi_i(\textbf{X};\mathcal{W}_i(t))$
   \STATE  $K_i = \mathcal{K}(A_i(\textbf{X}),A_i(\textbf{X}))$
   \STATE Update $\mathcal{W}_i(t)$ using SGD for loss in equation~\eqref{eqn:homo2}  
   \ENDFOR
   \STATE Return $\mathcal{W}_i(t)$ to the server 
\end{algorithmic}
\end{algorithm}

\begin{algorithm}[tb]
   \caption{FedHeNN Algorithm for Heterogeneous clients}
   \label{alg:fednn_algo_hetero}
   \begin{algorithmic}
   \STATE {\bfseries Input:} number of clients $N$, number of communication rounds $T$, 
   number of local epochs $E$, parameter $\eta_0$, weight vector for clients $[w_1,w_2,\dots w_N]$ 
   \STATE {\bfseries Output:} Final set of personalised models {$\mathcal{W}_1(T), \mathcal{W}_2(T) \dots \mathcal{W}_N(T)$} \\
   
   \STATE {\bfseries At Server - }
   \STATE Initialize {$\mathcal{W}_1(0), \mathcal{W}_2(0) \dots \mathcal{W}_N(0)$}
   \FOR{$t=1$ {\bfseries to} $T$}
    \STATE $\eta = \eta_0 \times f(t)$
    \STATE Generate RAD, \textbf{X}, by random sampling
    \FOR{each client $j$}
    \STATE $A_j(\textbf{X}) = \Phi_j(\textbf{X};\mathcal{W}_j(t-1))$
    \STATE $K_j = \mathcal{K}(A_j(\textbf{X}),A_j(\textbf{X}))$
    \ENDFOR
    \STATE $\Bar{K}(t-1)  =  \sum_{j=1}^{N} w_j K_j$
    \STATE Select a subset of clients $\mathcal{N}_t$  
    \FOR{each selected client $i \in \mathcal{N}_t$}
    \STATE $\mathcal{W}_i(t) =$ \textbf{LocalTraining}$( \mathcal{W}_i(t-1), \Bar{K}(t-1), \textbf{X}, \eta)$
   \ENDFOR
   \ENDFOR
   \STATE Return {$\mathcal{W}_1(T), \mathcal{W}_2(T) \dots \mathcal{W}_N(T)$}
   \STATE {\bfseries LocalTraining}$(\mathcal{W}_i(t-1), \Bar{K}(t-1), \textbf{X}, \eta)$
   \STATE Initialize $\mathcal{W}_i(t)$ with $(\mathcal{W}_i(t-1))$
   \FOR{each local epoch}
   \STATE  $A_i(\textbf{X}) = \Phi_i(\textbf{X};\mathcal{W}_i(t))$
   \STATE  $K_i = \mathcal{K}(A_i(\textbf{X}),A_i(\textbf{X}))$
   \STATE Update $\mathcal{W}_i(t)$ using SGD for loss in equation~\eqref{eqn:hetero2}
   \ENDFOR
   \STATE Return $\mathcal{W}_i(t)$ to the server 
\end{algorithmic}
\end{algorithm}

\vspace{-2ex}
\section{Experiments} \label{experiments}
We now present the effectiveness of the FedHeNN framework using empirical results on different datasets and models. We simulate statistically heterogeneous and system heterogeneous FL settings by manipulating the data partitions and model architectures across clients. We also discuss the effects of other variables like the size of RAD, $\eta$ and the choice of kernel on the performance of FedHeNN.

\subsection{Experimental Details}
We evaluate FedHeNN in different settings involving varied models, tasks, heterogeneity levels, and datasets. We start with descriptions of these settings followed by our results.

\textbf{Datasets} We consider two different high level tasks, image classification and text classification, and use datasets corresponding to these from the popular federated learning benchmark LEAF \cite{caldas2019leaf}. For the image classification task, we use CIFAR-10 and CIFAR-100 datasets that contain colored images in 10 and 100 classes respectively. And for the text classification task we use a binary classification dataset called Sentiment140. We partition the entire data to generate non-iid samples on each client and then split those into training and test sets at the client site. 

\textbf{Baselines} We compare our method against three different baselines - FedAvg, FedProx and FedRep. The FedAvg and FedProx algorithms learn a centralised global model thus the reported performance metric is of the global model. On the contrary, the FedRep method learns personalised models for each client. Therefore, FedRep's performance is reported for the personalised models. For FedHeNN, we report the performance of local models for both homogeneous and heterogeneous settings and that of the global model for the homogeneous setting. As for FedProto, it is demonstrated in the paper that the performance gap between FedProto and FedAvg decreases when we have more samples per class. In our settings, since we do not restrict the number of samples per class and also beat the FedAvg algorithm by a significant margin, we do not directly compare with FedProto.

\textbf{Evaluation Metric and other Parameters} We use the average test accuracy obtained on the clients' test datasets as the evaluation criterion. For global models the test accuracy is computed by evaluating the global model on the local test datasets and for the local models test accuracy is computed by testing personalised models on the local test datasets. For
hyperparameter tuning of methods, we utilise a global val-
idation dataset which is not shared with any of the clients. 

The hyperparameter $\eta$ that controls the contribution of representation similarity in the objective function is kept as a function of $t$ (the number of communication round). This is because the initial representations obtained from insufficiently trained models are not accurate and keeping a high $\eta$ in the initial rounds may mislead the training. The base value of $\eta_0$ is tuned as a hyperparameter and the best performance is obtained by keeping $\eta_0 = 0.001$ for CIFAR-10 and CIFAR-100, and $\eta_0 = 0.01$ for Sentiment140.

The size of RAD is an important parameter. The reported performance is obtained by keeping this size constant at 5000 which is much smaller than the size of training or test sets and doesn't increase the memory footprint drastically.  

\begin{table*}[ht]
\caption{Average test accuracy of FedHeNN computed for the common global model as compared to the baselines with global models.}
\label{test_accuracy_global}
\vskip 0.15in
\begin{center}
\begin{small}
\begin{sc}
\scalebox{0.9}{
\begin{tabular}{lccr}
\toprule
Data set(Setting) & FedAvg & FedProx & FedHeNN Global \\
\midrule
CIFAR-10(100 clients, 2 cls/client) & 44.29 $\pm$ 0.5 & 53.8 $\pm$ 2.3 & \textbf{68.8 $\pm$ 2.1} \\
CIFAR-10(100 clients, 5 cls/client) & 58.14 $\pm$ 0.7 & 63.3 $\pm$ 2.0 & \textbf{70.19 $\pm$ 2.0} \\
CIFAR-10(500 clients, 2 cls/client) & 42.7 $\pm$ 0.4 & 50.46 $\pm$ 1.4  & \textbf{65.4 $\pm$ 0.8} \\
CIFAR-10(500 clients, 5 cls/client) & 56.8 $\pm$ 0.5 & 55.2 $\pm$ 1.2   & \textbf{64.7 $\pm$ 0.7} \\
CIFAR-100(100 clients, 20 cls/client) & 28.6 $\pm$ 0.8 & 27.3 $\pm$ 1.1  & \textbf{44.2 $\pm$ 0.7} \\
Sentiment140(100 clients, 2 cls/client) & 52.6 $\pm$ 0.4 & 52.7 $\pm$ 1.0  & \textbf{52.7 $\pm$ 0.01} \\
\bottomrule
\end{tabular}}
\end{sc}
\end{small}
\end{center}
\vskip -0.1in
\end{table*}

\begin{table*}[ht]
\caption{Average test accuracy of FedHeNN computed for the personalised models as  compared to the baselines with personalised models.}
\label{test_accuracy_local}
\vskip 0.15in
\begin{center}
\begin{small}
\begin{sc}
\scalebox{0.9}{
\begin{tabular}{lccr}
\toprule
Data set(Setting) &  FedRep &  FedHeNN Homo & FedHeNN Hetero\\
\midrule
CIFAR-10(100 clients, 2 cls/client) & 85.7 $\pm$ 0.4  & 94.7 $\pm$ 1.1 & \textbf{88.9 $\pm$ 0.35}\\
CIFAR-10(100 clients, 5 cls/client) & 72.4 $\pm$ 1.2  &  84.37 $\pm$ 1.5 & \textbf{73.01 $\pm$ 0.3}\\
CIFAR-10(500 clients, 2 cls/client) & 78.9 $\pm$ 0.6 &  86.5 $\pm$ 0.9 & \textbf{82.02 $\pm$ 0.8}\\
CIFAR-10(500 clients, 5 cls/client) & 58.14 $\pm$ 0.21  & 73.32 $\pm$ 1.23 & \textbf{61.74 $\pm$ 0.6}\\
CIFAR-100(100 clients, 20 cls/client) & 38.85 $\pm$ 0.9 & 62.89 $\pm$ 0.8 & \textbf{43.36 $\pm$ 0.2}\\
Sentiment140(100 clients, 2 cls/client) & 69.8 $\pm$ 0.4 & \textbf{72.6 $\pm$ 0.3} & 71.5 $\pm$ 0.5\\
\bottomrule
\end{tabular}}
\end{sc}
\end{small}
\end{center}
\vskip -0.1in
\end{table*}

\textbf{Implementation} For the FL simulations, we keep the non-iid data distribution across clients such that each client will have access to data of only certain classes, for example, with 5 classes per client, $i^{th}$ client might have access to data for classes $\{2,3,4,5,8\}$ and client $j$ might have $\{1,4,5,6,7\}$. We also vary this number of classes to change the heterogeneity level across clients. The robustness and scalabality of our method is tested by increasing the number of clients participating in training from 100 to 500 for CIFAR-10 dataset. The total number of communication rounds is kept constant at 200 for all algorithms and at each round only 10\% of the clients are sampled and updated. We find that increasing the number of local epochs on clients doesn't worsen the performance of the client models for our method, so the number of local epochs is set to 20. For the heterogeneous FedHeNN, each entry of the weight vector for aggregating the representations $\textbf{w}$ is set to $\frac{1}{N}$. In each local update, we use SGD with momentum for training.  

For the homogeneous FedHeNN for CIFAR datasets, we use a CNN model with 2 convolutional layers with each convolutional layer followed by a max-pooling layer followed by 3 fully-connected layers at the end. For the heterogeneous FedHeNN, for each client we uniformly randomly sample from a set of 5 different CNNs obtained by varying the architecture size in between the simplest one that contains 1 convolutional and 1 fully connected layer and the most complex one containing 3 convolutional and 3 fully connected layers. For the Sentiment140 dataset, we use either a 1-layer or a 2-layer LSTM followed by 2 fully connected layers.

\textbf{CKA} For the CKA distance metric, we evaluate the performances by using a linear kernel as well as an RBF kernel. For the RBF kernel, we try various values for $\sigma$ but as we will show later, the performance obtained using RBF kernel is not very different from that of the linear kernel.

\subsection{Results}
The performance of FedHeNN and baselines under various settings is reported in Table~\ref{test_accuracy_global} and Table~\ref{test_accuracy_local}. The global model performances of the FedHeNN global model and related baselines is reported in Table~\ref{test_accuracy_global}. We observe that the FedHeNN global model outperforms the FedAvg and FedProx algorithms. The results for comparisons of the personalised models are reported in Table~\ref{test_accuracy_local} and the results demonstrate that the FedHeNN's performance is better than the baselines. We observe that the homogeneous FedHeNN has a higher gain over the baselines than the heterogeneous FedHeNN, which is expected because of the varying capacity of the local models in the heterogeneous setting.

\textbf{Linear vs RBF Kernel} For computing the CKA based distances, we try using both the Linear as well as RBF kernel. Based on the empirical analysis of FedHeNN, it is observed that both the linear and RBF kernels give comparable performances as shown in Table \ref{linear_rbf}.

\begin{table}[t]
\caption{Test Accuracy for Linear vs RBF Kernel compared for homogeneous FedHeNN on CIFAR-10 dataset.}
\label{linear_rbf}
\vskip 0.15in
\begin{center}
\begin{small}
\begin{sc}
\begin{tabular}{lcr}
\toprule
Dataset & Linear CKA & RBF CKA \\
\midrule
CIFAR-10 & 94.47 &  93.03 \\
CIFAR-100 & 84.37 & 83.03 \\
Sentiment140 & 72.6 & 72.8 \\
\bottomrule
\end{tabular}
\end{sc}
\end{small}
\end{center}
\vskip -0.1in
\end{table}

\textbf{Effect of Local epochs} We also analyse the effect of varying local epochs in FedHeNN. In FedAvg, increasing the number of local epochs has an adverse effect on the performance of the model. On the other hand, no such effect was observed for FedHeNN owing to the presence of the proximal term. We keep the number of local epochs for FedHeNN to be as high as 20.

\begin{figure}[ht]
\vskip 0.2in
\begin{center}
\centerline{\includegraphics[width=\columnwidth, height =5cm, width = 7cm]{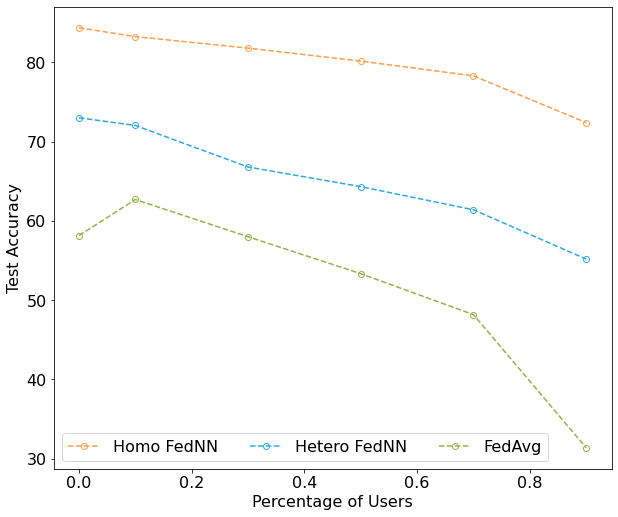}}
\caption{Changes in test accuracy when the data on a fraction of clients is reduced by 50\% shown for CIFAR-10 dataset.}
\label{change_data}
\end{center}
\vskip -0.2in
\end{figure}

\textbf{Sensitivity to Changing Amount of Data} We have empirically shown that FedHeNN is able to accommodate the clients with lower compute resources in an effective way. In order to show that the FedHeNN can also work with the clients with smaller data footprint, we do an experiment in which we randomly take a fraction of clients and reduce the data on those clients by 50\%. We show the results of the experiment in Figure \ref{change_data} where x-axis has the fraction of clients picked for shrinking the data and y-axis is the average test accuracy of all clients obtained when the framework is trained on the reduced dataset. We notice that even with the decreasing data size on the clients' ends FedHeNN is able to maintain graceful performance. This effect could be attributed to the fact that even though the number of instances to train the prediction function is reduced, the representation learning component is still robust.

\section{Discussion} \label{discussion}
In this paper, we present a new method for enhancing learning in Federated Learning by introducing a systematic framework called FedHeNN. The FedHeNN framework is unique because it allows the clients with heterogeneous architectures to participate in the joint learning process helping boost the performance. This could be a huge practical advancement as now the individual client devices or organisations with variable amount of resources can equally contribute and learn from each other. The empirical results indicate that FedHeNN is able to achieve better performing results while also being more inclusive. For future work, we would work on determining the solution characteristics and the convergence guarantees of both the FedHeNN algorithms. There are a few natural research directions arising from this work. First, it is of interest to extend the FedHeNN to the Transformer architectures~\cite{vaswani2017attention, Ho_Transformermixture, Ho_Fourier}. Second, we can use generative models to generate the RAD, the data being used for aligning representations to relax the assumption of presence of additional data on the server.

\section*{Acknowledgements}
This work was supported by Office of Naval Research(ONR) [N00014-19-1-2625]. We thank Dr. Ravi Srinivasan and Dr. Alex Liu for their valuable suggestions. 
NH gratefully acknowledges support from the NSF [IFML 2019844] award and research gifts by the NSF AI Institute for Foundations of Machine Learning. 

\bibliography{refs}
\bibliographystyle{icml2022}
\end{document}